\documentclass[11pt]{article}

\usepackage[preprint]{acl}

\usepackage{times}
\usepackage{latexsym}
\usepackage[T1]{fontenc}
\usepackage[utf8]{inputenc}
\usepackage{microtype}
\usepackage{inconsolata}
\usepackage{graphicx}
\usepackage{amsmath}
\usepackage{amssymb}
\usepackage{amsthm}
\usepackage{booktabs}
\usepackage{multirow}
\usepackage{algorithm}
\usepackage{algpseudocode}
\algnewcommand\algorithmicinitialization{\textbf{Initialization:}}
\algnewcommand\Initialization{\item[\algorithmicinitialization]}
\usepackage{xcolor}
\usepackage[most]{tcolorbox}
\usepackage{enumitem}

\newtheorem{definition}{Definition}

\title{Unlocking the Unsolvable: \\
       Teacher-Guided Curriculum for Data-Efficient RLVR}

\author{
Yukang Zhu \\
Independent Researcher \\
\texttt{zhuyukang2000@gmail.com}
\And
Zhen Han\thanks{This work was conducted independently of Amazon and does not relate to the author's position there.} \\
Amazon \\
\texttt{zhenhz@amazon.com}
}

\begin{document}
\maketitle

\begin{abstract}
Reinforcement Learning with Verifiable Rewards (RLVR) has shown remarkable
success in improving the mathematical reasoning of large language models.
Yet problems beyond the model's current capability, where rollouts uniformly fail
and no learning signal is produced, are structurally wasted despite marking
the most informative training frontier.
We show that these otherwise-inert problems can be \emph{unlocked} via teacher-guided
curriculum learning: partial reasoning traces from a stronger model create a
graded difficulty landscape, and a backward-chaining curriculum progressively
withdraws guidance until the model solves problems unaided.
Training on only 128 unsolvable problems matches or exceeds GRPO trained on
a full 2{,}000-problem corpus ($\sim$16$\times$ data efficiency) on the
nine-benchmark average for both base models, while substantially expanding
the reasoning boundary measured by pass@$k$ at large $k$.
Furthermore, we identify a distribution-shift cost that is particularly
acute in the unsolvable-only regime and propose \textbf{Monotone Frontier
Curriculum (MFC)}, a method that monotonically
drives training toward unguided solving, consistently outperforming
existing curriculum methods.
\end{abstract}

\section{Introduction}
\label{sec:intro}

\begin{figure*}[t!]
\centering
\includegraphics[width=\textwidth]{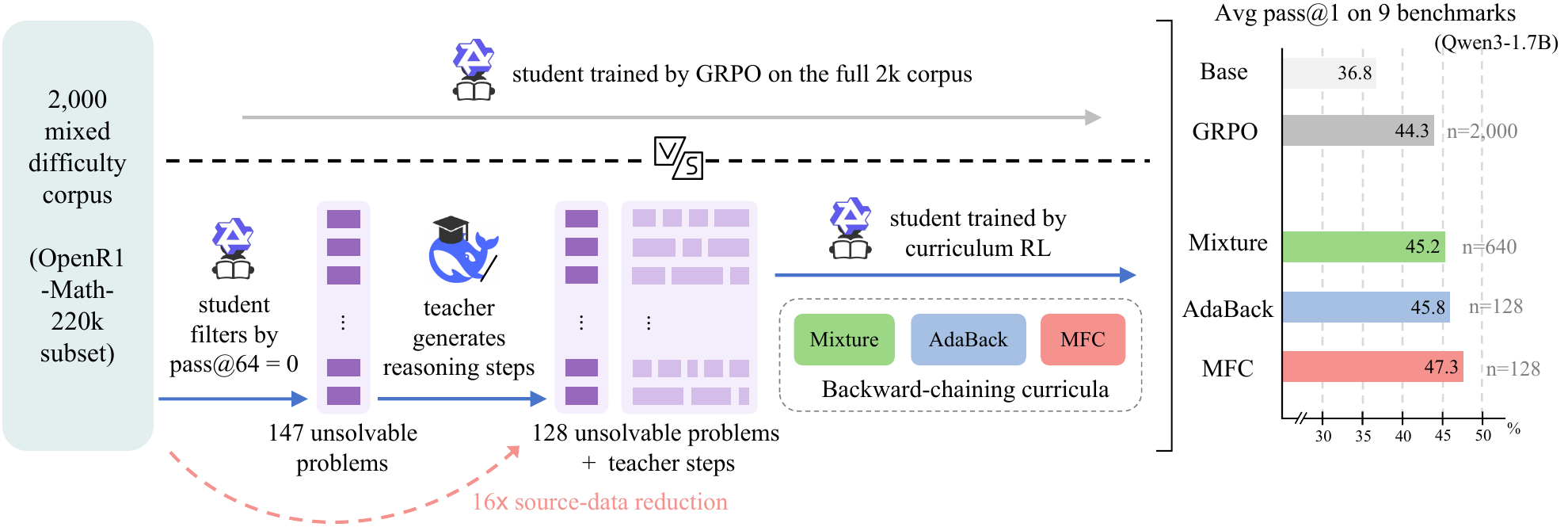}
\caption{\textbf{Overview of our unsolvable-curriculum pipeline.}
Top path: standard GRPO trained on the full $2{,}000$-problem mixed-difficulty corpus.
Bottom path: we filter the same corpus to its $147$ unsolvable problems ($\mathrm{pass}@64 = 0$ on the student), generate a step-structured trace with a stronger teacher, discard teacher failures ($136$ remain), and select the $128$ shortest examples as the curriculum-training set.
We compare three backward-chaining curricula: Mixture (R$^3$), AdaBack, and our \textbf{MFC}.
MFC is designed for the unsolvable-only regime.
With $\sim$16$\times$ less source data, the curriculum methods match or exceed the GRPO baseline on the cross-benchmark average (right; Qwen3-1.7B, $9$ benchmarks).}
\label{fig:overview}
\end{figure*}

Reasoning-centric large language models (LLMs) \cite{o1,DeepSeek-R1} have established Reinforcement Learning with Verifiable Rewards (RLVR) \cite{GRPO} as a powerful recipe for improving mathematical reasoning.
RLVR builds on the broader paradigm of reinforcement learning from human feedback \cite{RLHF} and policy gradient methods \cite{PPO}, replacing costly human annotations with verifiable reward signals.
While most work focuses on algorithmic refinements \cite{DAPO, Dr.GRPO, PRIME}, a complementary question has received growing attention: \emph{which training data matters most}?
\citet{Hard_Examples} show that the hardest problems yield the largest gains, and \citet{LILO} show that expected policy improvement is bounded by the variance of success.
\citet{1-shot_RLVR} demonstrate that even a single example can unlock substantial reasoning gains.
These findings suggest a common theme: problems near the model's reliable-solving boundary, with mixed rollout outcomes, tend to provide the richest learning signal for RLVR.

A separate line of inquiry concerns the \emph{scope} of what RLVR can teach. 
\citet{Yue_et_al.} find that standard RLVR raises $\mathrm{pass}@1$ by amplifying rewarded traces already in the base model, while at large $k$ the base model matches or exceeds the RL-trained model and coverage can shrink.
\citet{Havrilla_et_al.} find that RL fine-tuning fails to explore far beyond solutions already available after supervised fine-tuning.
The model's \emph{reasoning boundary}\footnote{Following \citet{Yue_et_al.}, we use ``reasoning boundary'' to denote the set of problems a model can solve given sufficient sampling, operationalized as $\mathrm{pass}@k$ at large $k$.} therefore tends to plateau or even shrink after standard training.
\citet{yao2025debaterlvrreasoningcapability} treat expansion and shrinkage as two stages of probability-mass reallocation: early exploitation can narrow coverage, and expansion requires training that continues into a later exploration stage.

What, then, about problems \emph{operationally beyond} that boundary?
In challenging domains, a non-trivial fraction of problems are \emph{unsolvable}: every sampled rollout fails even under generous budgets (\emph{e.g.}, $\mathrm{pass}@64 = 0$).
We find ${\sim}7\%$ of a 2{,}000-problem competition-math corpus is unsolvable for Qwen3-1.7B.
These problems present a paradox: they mark the model's reasoning boundary and carry significant learning potential, yet under standard GRPO, uniform failure produces vanishing advantage and near-zero gradient.\par
\pagebreak[4]
\noindent This motivates our central question:
\begin{quote}
\emph{Can unsolvable problems be \textbf{unlocked}\footnote{Throughout this paper, \emph{unlocking} refers to converting a problem that produces vanishing gradient under RLVR (due to uniform rollout failure) into one that yields non-trivial learning signal and meaningful group-relative advantage.} for RLVR, and can they alone drive data-efficient reasoning improvement and frontier expansion?}
\end{quote}

We answer affirmatively. 
Our key insight is that a partial reasoning trace from a stronger teacher transforms a single unsolvable problem into a \emph{family of states} with graded difficulty, indexed by a guidance level $\rho \in [0,1]$ controlling how much of the teacher's reasoning is revealed. 
A backward-chaining curriculum over $\rho$ can then be utilized to progressively withdraw guidance toward unaided solving.

Figure~\ref{fig:overview} summarizes the pipeline. Training on \textbf{128 unsolvable problems} with backward-chaining curricula \cite{AdaBack} matches or exceeds GRPO on the full 2{,}000-problem corpus (${\sim}16\times$ data reduction) on the nine-benchmark average for both base models, and \emph{substantially expands} the reasoning boundary (AIME $\mathrm{pass}@k$ improves at every $k$ up to $256$).

We further observe that existing backward-chaining curricula are not optimized for the unsolvable-only regime.
Every training step at guidance level $\rho > 0$ occurs under a distribution that differs from the test-time distribution ($\rho = 0$), and generic curricula have no explicit mechanism to maximize training at $\rho = 0$.
We formalize this distribution-shift cost in \S\ref{sec:dist-shift} and propose \textbf{Monotone Frontier Curriculum (MFC)}, a method that maintains a per-sample monotone non-increasing guidance frontier and consistently outperforms existing curricula.

Our contributions are as follows:
\begin{itemize}
  \item We demonstrate that \textbf{teacher-guided curriculum learning unlocks unsolvable problems for RLVR}. Training on 128 such problems achieves data efficiency comparable to or exceeding GRPO on 2{,}000 mixed-difficulty problems.
  \item We show that this paradigm \textbf{expands the reasoning boundary} ($\mathrm{pass}@k$ gains at every $k$ up to 256), contrasting with prior observations that RLVR narrows coverage \cite{Yue_et_al.}.
  \item We propose \textbf{Monotone Frontier Curriculum (MFC)}, designed for the unsolvable-only regime, which maximizes unguided training through a monotone per-sample frontier and consistently outperforms existing curricula.
\end{itemize}

\section{The Unsolvable Regime}
\label{sec:prelim}

We first establish the empirical foundation for our work: we formalize what it means for a problem to be \emph{unsolvable} under RLVR (\S\ref{sec:unsolvable-def}), verify that most of such problems remain inert throughout training (\S\ref{sec:waste}), and show that teacher guidance can unlock them as a rich source of learning signal (\S\ref{sec:unlock}).
\subsection{Unsolvable Problems and Vanishing Gradient Signal}
\label{sec:unsolvable-def}

GRPO \cite{GRPO} trains a policy $\pi_\theta$ by sampling $n$ rollouts per prompt and computing a group-relative advantage $A_i = (r_i - \mu_g) / (\sigma_g + \epsilon)$. When every rollout returns the same reward, $\sigma_g = 0$ and all advantages vanish, producing no gradient. For binary rewards this occurs whenever a group is uniformly correct or uniformly incorrect.

\begin{definition}[Unsolvable problem]
\label{def:unsolvable}
For a policy $\pi_0$ and a sampling budget $N$, a problem $s$ is \emph{$N$-unsolvable} if $\mathrm{pass}@N(\pi_0, s) = 0$, i.e., no correct solution is found among $N$ independent rollouts.
\end{definition}

The budget $N$ should be large enough relative to the GRPO per-prompt rollout count $n$ so that $N$-unsolvability reliably predicts uniform failure within each training group.
We use $n = 8$ rollouts per group and set $N = 64$: a problem with no success in 64 attempts is overwhelmingly likely to produce $n = 8$ failures at almost every training step. From our 2{,}000-problem competition-mathematics corpus (details in Appendix~\ref{sec:appendix-data}), approximately 7\% of problems ($147$ total) are 64-unsolvable for Qwen3-1.7B.
These problems potentially mark the model's reasoning boundary, yet they contribute negligible gradient signal under GRPO.
\subsection{The Waste Hypothesis}
\label{sec:waste}

One might expect training to gradually render some unsolvable problems productive, so that they begin generating learning signals. 
We test this by tracking the $147$ unsolvable problems across all checkpoints of a 1{,}200-step GRPO baseline on Qwen3-1.7B (the same run described in \S\ref{sec:experiments}), evaluating $\mathrm{pass}@8$ at each checkpoint.

\begin{figure}[t]
\centering
\includegraphics[width=\columnwidth]{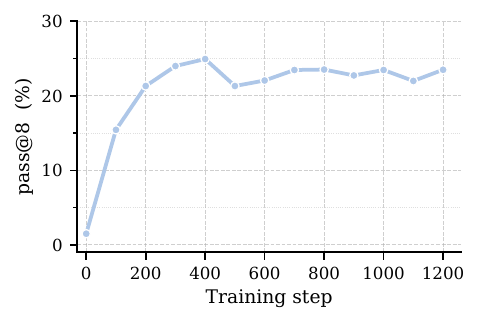}
\caption{Average $\mathrm{pass}@8$ on the $147$ initially unsolvable problems ($\mathrm{pass}@64 = 0$) across GRPO training checkpoints.}
\label{fig:waste}
\end{figure}

As shown in Figure~\ref{fig:waste}, $\mathrm{pass}@8$ rises to $\sim$24\% by step 300--400 then plateaus; at step 1{,}200, more than 75\% of initially unsolvable problems still yield zero successes. 
The small fraction that becomes learnable can be primarily explained by the model's modest capability growth during training.
This confirms the \textbf{waste hypothesis}: under GRPO, the vast majority of unsolvable problems are \emph{structurally excluded} from learning throughout training.

\subsection{Unlocking via Teacher Guidance}
\label{sec:unlock}

If the bottleneck is uniform failure, a natural remedy is to reduce effective difficulty so that some succeed.
We employ DeepSeek-V3.2 \cite{DeepSeek-V3.2,DeepSeek-V3} as the teacher model throughout this work; 
for each unsolvable problem $s$, we prompt the teacher model to generate a structured reasoning trace $T(s)$ consisting of discrete steps (details in Appendix~\ref{sec:prompt-templates}). 
A \emph{guidance level} $\rho \in [0, 1]$, then, controls the fraction of this trace revealed to the student. 
Letting $n_\mathrm{steps}(s)$ denote the number of steps in $T(s)$, we expose the first $g$ steps:

\begin{equation}
\begin{aligned}
  g &= \mathrm{round}(\rho \cdot n_\mathrm{steps}(s)), \\
  s_\rho &= s \,\cup\, \{\text{first $g$ steps of $T(s)$}\}.
\end{aligned}
\label{eq:rho-lattice}
\end{equation}
The student receives the augmented prompt $s_\rho$ in which the teacher's partial trace is presented as auxiliary context (a ``hint'') and then generates a complete solution from scratch (see Appendix~\ref{sec:prompt-templates} for templates).

Table~\ref{tab:unlock-curve} shows the \emph{unlock curve}: $\mathrm{pass}@8$ as a function of $\rho$ on base Qwen3-1.7B over the $147$ unsolvable problems. We measure $\mathrm{pass}@8$ because it mirrors the GRPO rollout regime ($n=8$): a problem with $\mathrm{pass}@8 > 0$ yields non-trivial group-relative advantage.

\begin{table}[t]
\centering
\small
\setlength{\tabcolsep}{5pt}
\begin{tabular}{lccccc}
\toprule
\textbf{Guidance $\rho$} & 0 & 0.25 & 0.50 & 0.75 & 1.00 \\
\midrule
$\mathrm{pass}@8$ & 2.2\% & 34.0\% & 53.6\% & 72.3\% & 85.4\% \\
\bottomrule
\end{tabular}
\caption{$\mathrm{pass}@8$ of Qwen3-1.7B on the $147$ unsolvable problems at varying guidance levels $\rho$ (averaged over $4$ independent evaluations).}
\label{tab:unlock-curve}
\end{table}

The curve rises sharply.
$\mathrm{pass}@8$ increases from $2.2\%$ at $\rho = 0$ to $34.0\%$ at $\rho = 0.25$ and $85.4\%$ at $\rho = 1$, with a broad learnable band in $\rho \in [0.25, 1]$.
The $2.2\%$ residual means that the per-problem success probability is small but not strictly zero, which remains consistent with the finite-sample $\mathrm{pass}@64 = 0$ filter.
Teacher guidance converts the otherwise inert population into one that produces non-trivial advantage at every step, the operational meaning of \emph{unlocking}.

Each unsolvable problem thus becomes a \emph{family of training states} indexed by $\rho$, naturally suggesting a \emph{backward-chaining curriculum}: 
begin at high $\rho$ where the student reliably succeeds, and progressively withdraw guidance toward $\rho = 0$, where the student must solve the problem entirely on its own.
We formalize this paradigm in the next section.

\section{Curriculum Methods for Unsolvable Problems}
\label{sec:methods}

\subsection{The Backward-Chaining Family}
\label{sec:bc-family}

A curriculum is \emph{backward-chaining} if it places non-trivial training mass at $\rho > 0$ and progressively shifts that mass toward $\rho = 0$. 
Three canonical sub-families occupy this design space.

\paragraph{Staged curricula.}
A staged curriculum maintains a single global schedule $\rho_\mathrm{global}(t)$ advanced when an aggregate success threshold is met \cite{Salimans_&_Chen_2018}.
In practice, curriculum performance is dominated by scheduler design, with no settled best practice for when and how to advance the global $\rho$. Vanilla staged reverse curricula can exhibit sharp drops at stage transitions and degradation of earlier-stage skill \cite{R3}.
For the experiments of \S\ref{sec:experiments} we therefore do not include a staged baseline.

\paragraph{Mixture curricula.}
A mixture curriculum bypasses the scheduler by expanding the per-sample state space: each problem is replicated at $M$ discretized $\rho$-levels, and a single training pass uniformly samples across all levels.
For the experiments of \S\ref{sec:experiments} we use R$^3$ \cite{R3} as the Mixture baseline with $M = 5$ levels ($\rho \in \{0, 0.25, 0.5, 0.75, 1\}$), expanding the $128$ unsolvable problems into a $640$-row dataset trained with standard GRPO.

\paragraph{Per-sample adaptive curricula.}
A per-sample adaptive curriculum maintains one schedule per sample and updates it based on rollout outcomes. 
The reference instantiation is AdaBack \cite{AdaBack}, a per-sample adaptive method that we compare with the R$^3$-style mixture below.
Originally proposed over a continuous guidance level, AdaBack is adapted here to share MFC's (\S\ref{sec:mfc}) discrete step lattice,\footnote{From here on we operate on the integer $g \in \{0, \ldots, n_\mathrm{steps}(s)\}$ of Eq.~\eqref{eq:rho-lattice}; the continuous $\rho$ is retained only for narrative description.} ensuring both methods sample from the same space (Appendix~\ref{sec:impl}).
Per sample $s$, AdaBack maintains an integer interval $[g_\mathrm{min}(s), g_\mathrm{max}(s)] \subseteq \{0, \ldots, n_\mathrm{steps}(s)\}$ initialized to the full range.
The first query uses $\rho = 0.5$, and subsequent visits sample $g_\mathrm{used} \sim \mathrm{Uniform}\{g_\mathrm{min}, \ldots, g_\mathrm{max}\}$.
AdaBack then updates the interval \emph{bidirectionally}: a success ($\bar r \geq \tau$) sets $g_\mathrm{max} \leftarrow g_\mathrm{used}$ and resets $g_\mathrm{min} \leftarrow 0$; a failure ($\bar r < \tau$) raises $g_\mathrm{min} \leftarrow g_\mathrm{used}$.
This implements a binary search for the per-sample \emph{learnable band} around which AdaBack stabilizes.
\subsection{The Hidden Cost in the Unsolvable-Only Regime}
\label{sec:dist-shift}

The unsolvable-only regime exposes a hidden inefficiency shared by all three methods above.
We make this precise by separating the curriculum's training-time objective from the test-time objective it ultimately serves.

The training objective at step $t$ under curriculum $\mathcal{C}$ samples $(s, \rho) \sim \mathcal{C}^{(t)}$ and conditions the policy on the augmented prompt $s_\rho$ of Eq.~\eqref{eq:rho-lattice}:
\begin{equation}
  J_\mathcal{C}^{(t)}(\theta) \;=\;
  \mathbb{E}_{(s, \rho) \sim \mathcal{C}^{(t)}}\,
  \mathbb{E}_{y \sim \pi_\theta(\cdot \mid s_\rho)}\bigl[r(s, y)\bigr],
  \label{eq:train-obj}
\end{equation}
where $r(s, y)$ is based on the \emph{original} problem $s$ but the policy sees $s_\rho$ whenever $\rho > 0$. The test-time objective is the unguided case
\begin{equation}
  J_0(\theta) \;=\; \mathbb{E}_{s \sim \mathcal{D}}\,
  \mathbb{E}_{y \sim \pi_\theta(\cdot \mid s)}\bigl[r(s, y)\bigr].
  \label{eq:test-obj}
\end{equation}
Whenever $g(s, \rho) > 0$, the curriculum induces a \emph{distribution shift} between training and test prompts. Improving $J_\mathcal{C}^{(t)}$ does not automatically improve $J_0$: augmented visits transfer to the unguided regime through parameter sharing, but the transfer is sign-indeterminate in general (Appendix~\ref{sec:appendix-gradient}).
A curriculum that aims to improve $J_0$ should therefore allocate training mass to the unguided regime.

We capture this via the \textbf{unguided training mass}
\begin{equation}
  M_0^{(t)} \;:=\; \Pr_{(s, \rho) \sim \mathcal{C}^{(t)}}\bigl[g(s, \rho) = 0\bigr],
  \label{eq:m0-def}
\end{equation}
the in-batch fraction of visits at guidance length zero. Higher $M_0^{(t)}$ means more compute on prompts matching the test distribution. 
Appendix~\ref{sec:appendix-gradient} provides the formal first-order decomposition of $J_0$ motivating this choice.

\begin{figure}[t]
\centering
\includegraphics[width=\columnwidth]{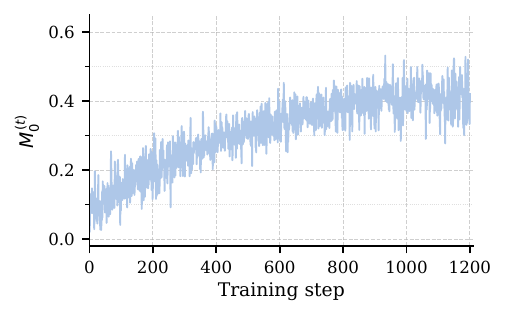}
\caption{Unguided training mass $M_0^{(t)}$ of AdaBack on the $128$-problem unsolvable corpus (Qwen3-1.7B, the same run in \S\ref{sec:experiments}). $M_0^{(t)}$ rises initially but plateaus once per-sample intervals settle in the learnable band.}
\label{fig:adaback-m0}
\end{figure}

This surrogate matters more in the unsolvable-only regime than in mixed-difficulty training. 
In a mixed corpus, many samples are already feasible at $\rho = 0$, so $M_0^{(t)}$ rises naturally. 
In an unsolvable-only corpus, samples produce near-zero feasibility at $\rho = 0$ initially, and none of the existing curricula carry an explicit mechanism to drive descent toward $\rho = 0$ as the policy improves. 
AdaBack, for instance, binary-searches the per-sample \emph{learnable band}, which for an unsolvable problem typically remains above $\rho = 0$ throughout training even as the policy improves substantially, so $M_0^{(t)}$ plateaus.
Figure~\ref{fig:adaback-m0} confirms this empirically: AdaBack's $M_0^{(t)}$ rises to $\sim$0.40 by step $800$ and remains essentially flat over the remaining steps.

\subsection{Monotone Frontier Curriculum (MFC)}
\label{sec:mfc}

The analysis above identifies a need for a curriculum that can continue increasing its unguided training mass as the policy improves.
MFC implements this goal by lowering a sample's guidance frontier only after the student succeeds with less guidance.
As a result, each sample's unguided sampling probability increases monotonically with demonstrated competence, and sampled $\rho$ shifts toward $0$.
It restructures AdaBack's update around an asymmetric principle: \emph{success commits, failure is forgotten}.
AdaBack raises $g_\mathrm{min}$ on failure.
MFC keeps only the success-side ratchet, committing $g_\mathrm{curr} \leftarrow g_\mathrm{used}$ on a successful visit if smaller, while failures leave the per-sample state untouched and still contribute to the GRPO advantage.
With $g_\mathrm{min}$ removed, the state collapses to a single integer $g_\mathrm{curr}(s)$ that is monotonically non-increasing in $t$.

\begin{algorithm}[t]
  \caption{Monotone Frontier Curriculum (MFC)}
  \label{alg:mfc}
  \begin{algorithmic}[1]
  \Require Training corpus $\mathcal{D}$; for each prompt $s \in \mathcal{D}$, a teacher trace $T(s)$ given as an ordered list of $n_\mathrm{steps}(s)$ reasoning steps; success threshold $\tau$; rollouts per prompt $n$.
  \Initialization $g_\mathrm{curr}(s) \leftarrow n_\mathrm{steps}(s)$ for all $s \in \mathcal{D}$.
  \For{each visit to $s \in \mathcal{D}$ during GRPO training}
    \State $g_\mathrm{used} \sim \mathrm{Uniform}\{0, 1, \ldots, g_\mathrm{curr}(s)\}$
    \State Roll out $n$ trajectories from $s$ with the first $g_\mathrm{used}$ steps of $T(s)$ prepended as hint
    \State Compute mean reward $\bar r$ and apply the GRPO update
    \If{$\bar r \geq \tau$ \textbf{and} $g_\mathrm{used} < g_\mathrm{curr}(s)$}
      \State $g_\mathrm{curr}(s) \leftarrow g_\mathrm{used}$
    \EndIf
  \EndFor
  \end{algorithmic}
  \end{algorithm}

This asymmetry suits the unsolvable-only regime: with a rapidly evolving policy, failure at low $\rho$ today is weak evidence for failure tomorrow, whereas success is firm evidence that a lower frontier is feasible.
The monotone descent propagates directly to the surrogate: each successful visit at a lower $g_\mathrm{used}$ permanently raises MFC's per-sample on-target probability $1/(g_\mathrm{curr}(s)+1)$ via the discrete-uniform draw of Algorithm~\ref{alg:mfc}. Once $g_\mathrm{curr}(s) = 0$, all visits to that sample are at $\rho = 0$ (equivalently, $\rho_\mathrm{max}(s) := g_\mathrm{curr}(s) / n_\mathrm{steps}(s) = 0$).

Algorithm~\ref{alg:mfc} gives the full procedure with a single hyperparameter $\tau$ (success threshold); we use $\tau = 0.5$ throughout.

\section{Experiments}
\label{sec:experiments}

\begin{table*}[t]
\centering
\footnotesize
\setlength{\tabcolsep}{4pt}
\renewcommand{\arraystretch}{1.1}
\resizebox{\textwidth}{!}{%
\begin{tabular}{ll|cc|ccccc|cc|c}
\toprule[1.5pt]
 & & \multicolumn{2}{c|}{In-domain} & \multicolumn{5}{c|}{Math OOD} & \multicolumn{2}{c|}{Non-math OOD} & \\
\textbf{Base Model} & \textbf{Method}
  & \textbf{OR1-200} & \textbf{Uns-22} & \textbf{MATH-500} & \textbf{AMC23}
  & \textbf{AIME24} & \textbf{AIME25} & \textbf{AIME26}
  & \textbf{SciBench} & \textbf{GPQA-D}
  & \textbf{Avg.} \\
\midrule[0.75pt]
\multirow{5}{*}{\textbf{Qwen3-1.7B}}
  & \textsc{Base}                  & 47.5 & 0.6 & 83.5 & 62.8 & 23.2 & 22.6 & 22.1 & 34.1 & 34.9 & 36.8 \\
  & \textsc{GRPO}                  & \textbf{68.6} & 23.9 & \textbf{89.5} & \underline{73.8} & 26.7 & 26.3 & 21.3 & 34.5 & 33.8 & 44.3 \\
  & \textsc{Mixture}               & 66.9 & 25.6 & 88.2 & 72.2 & 29.8 & 25.8 & \underline{27.1} & \underline{36.3} & \underline{35.0} & 45.2 \\
  & \textsc{AdaBack}               & 67.2 & \underline{29.5} & \underline{89.1} & 72.5 & \underline{30.0} & \textbf{29.2} & 26.3 & 34.1 & 34.1 & \underline{45.8} \\
  & \textbf{\textsc{MFC}}          & \underline{68.2} & \textbf{31.8} & \underline{89.1} & \textbf{75.0} & \textbf{32.5} & \underline{28.8} &  \textbf{27.2} & \textbf{37.1} & \textbf{36.1} & \textbf{47.3} \\
\midrule[0.75pt]
\multirow{5}{*}{\textbf{Qwen3-0.6B}}
  & \textsc{Base}                  & 28.8 & 0.0  & 67.1 & 38.1 & 4.2 & 11.3 & 8.8 & 18.4 & 29.3 & 22.9 \\
  & \textsc{GRPO}                  & \textbf{51.1} & 16.5 & \textbf{78.8} & 50.0 & \underline{11.7} & 15.4 & 8.8 & 21.2 & 28.3 & 31.3 \\
  & \textsc{Mixture}               & 45.4 & 19.3 & 74.4 & 50.0 & 9.6 & 14.6 & 9.6 & 21.5 & 30.6 & 30.6 \\
  & \textsc{AdaBack}               & 49.0 & \underline{22.7} & 75.5 & \underline{51.3} & 10.8 & \underline{18.8} & \underline{12.5} & \textbf{23.0} & \underline{31.7} & \underline{32.8} \\
  & \textbf{\textsc{MFC}}          & \underline{50.4} & \textbf{23.3} & \underline{76.8} & \textbf{52.2} & \textbf{13.3} & \textbf{19.6} & \textbf{14.2} & \underline{22.8} & \textbf{33.1} & \textbf{34.0} \\
\bottomrule[1.5pt]
\end{tabular}
}
\caption{Main result. $\mathrm{pass}@1$ (\%, average of $8$ completions at temperature $1$) on nine benchmarks for two base models. \textbf{Bold} marks the column-wise best and \underline{underline} the column-wise second-best, both computed \emph{within} each base-model block.}
\label{tab:main}
\end{table*}

\subsection{Setup}
\label{sec:setup}

\paragraph{Base models.}
We use \textbf{Qwen3-1.7B} \cite{Qwen3} as the primary base model and \textbf{Qwen3-0.6B} as a smaller-scale replicate, both from the released base (non-instruct) checkpoint. Both models belong to the Qwen model family \cite{Qwen2.5,Qwen3} and use the same tokenizer.

\paragraph{Datasets.}
All training and in-domain evaluation data are subsets of OpenR1-Math-220k \cite{OpenR1-Math}, derived through the quality-filter process of Appendix~\ref{sec:appendix-data}. 
We use a $2{,}000$-problem mixed-difficulty training set (\textbf{OR1-2k}) and a disjoint $200$-problem held-out test set (\textbf{OR1-200}). 
The unsolvable training pool is built directly from OR1-2k: we evaluate Qwen3-1.7B at $\mathrm{pass}@64$ and retain the 147 problems with $\mathrm{pass}@64 = 0$, generate teacher traces with DeepSeek-V3.2 and discard problems on which the teacher itself fails (136 remaining), and select the $128$ shortest by combined question-plus-trace length to bound prompt budget, yielding \textbf{Uns-128}. 
The same Uns-128 is reused unchanged when training Qwen3-0.6B (a fortiori unsolvable for the weaker sibling). 
The analogous extraction on OR1-200 yields \textbf{Uns-22}, an in-domain unsolvable evaluation subset.
We release the full OR1-2k, OR1-200, Uns-128, and Uns-22 records, together with the Uns-128 teacher traces, in a public Hugging Face dataset repository.\footnote{\url{https://huggingface.co/datasets/yukangzhu/unlocking-the-unsolvable}}

\paragraph{Algorithms.}
For each base model we compare a \emph{Base} (no-RL) reference and four trained algorithms, all run for $1{,}200$ GRPO steps under matched compute: \textbf{(1) GRPO} on the full OR1-2k corpus with no teacher guidance; 
\textbf{(2) Mixture Curriculum} (R$^3$-style; \S\ref{sec:bc-family}) on Uns-128 expanded into a $640$-row dataset at $M = 5$ uniform $\rho$-levels; \textbf{(3) AdaBack} on Uns-128 with the discrete adaptation of \S\ref{sec:bc-family} ($\tau = 0.5$, default initial $\rho = 0.5$); 
and \textbf{(4) MFC} (ours) on Uns-128 with $\tau = 0.5$ (Algorithm~\ref{alg:mfc}). 
All teacher-guided runs provide the partial trace as a hint within the prompt (\S\ref{sec:unlock});
all four share an identical GRPO core (batch size $128$, $n=8$ rollouts at temperature $1$, $1\mathrm{e}{-6}$ learning rate, low-variance KL with coefficient $1\mathrm{e}{-3}$, $8{,}192$-token response cap) augmented with two DAPO \cite{DAPO} stabilizers: \emph{clip-higher} and an \emph{overlong-response penalty}. 
Full hyperparameter tables and launcher pointers are in Appendix~\ref{sec:impl}.

\paragraph{Benchmarks.}
Evaluation spans \textbf{nine} benchmarks in three groups: two \emph{in-domain math} sets (\textbf{OR1-200} and \textbf{Uns-22}, the latter directly probing in-domain reasoning-boundary lift); 
five \emph{out-of-distribution math} sets (MATH-500 \cite{MATH,MATH-500}, AMC23\footnote{\url{https://huggingface.co/datasets/zwhe99/amc23}}, AIME24/25/26\footnote{\url{https://huggingface.co/datasets/hendrydong/aime24}, \url{https://huggingface.co/datasets/math-ai/aime25}, \url{https://huggingface.co/datasets/math-ai/aime26}}); 
and two \emph{out-of-distribution non-math reasoning} sets (SciBench \cite{SciBench} and GPQA-Diamond \cite{GPQA}). 
All scores are $\mathrm{pass}@1$ averaged over $8$ completions at temperature $1$; correctness is judged by Math-Verify\footnote{\url{https://github.com/huggingface/Math-Verify}}, a rule-based verifier that normalizes and compares model outputs against ground-truth answers.

\begin{figure*}[t]
\centering
\includegraphics[width=\textwidth]{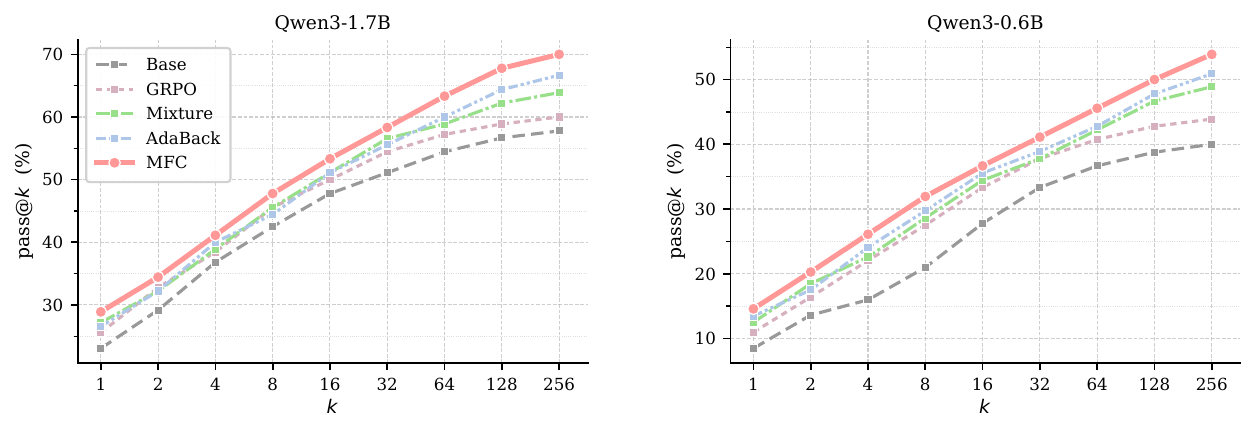}
\caption{Reasoning boundary measured by $\mathrm{pass}@k$ on the union of AIME24/25/26 ($90$ problems), $k \in \{1, 2, 4, 8, 16, 32, 64, 128, 256\}$ on a $\log_2$ axis. Left: Qwen3-1.7B; right: Qwen3-0.6B. The unsolvable-curriculum methods (Mixture, AdaBack, MFC) widen the gap over both \emph{Base} and GRPO at large $k$, with MFC uniformly best.}
\label{fig:boundary}
\end{figure*}

\subsection{Main Results}
\label{sec:main-results}

Table~\ref{tab:main} reports $\mathrm{pass}@1$ on all nine benchmarks for both base models and the four trained algorithms, with the un-trained \emph{Base} reference for context. 
We summarize three results from Table~\ref{tab:main}:

\begin{enumerate}
  \item \textbf{$128$ unsolvable problems match or exceed $2{,}000$ mixed.} On Qwen3-1.7B, AdaBack and MFC stay within $1.4$~pp of the GRPO~2k baseline on OR1-200 and MATH-500. On Qwen3-0.6B, the corresponding gaps range from $0.7$ to $3.3$~pp. Both methods exceed GRPO on Uns-22 and on the average across AIME24/25/26 (e.g.\ $+7.9$ / $+6.8$ pp for MFC on Uns-22), with a $\sim$16$\times$ cut in source data. Mixture is competitive on Qwen3-1.7B but trails GRPO on the Qwen3-0.6B average ($30.6$ vs $31.3$), so uniform-$\rho$ mixing alone is not enough at the smaller scale.
  \item \textbf{MFC has the best cross-benchmark average on both base models.} Its averages are $47.3$ and $34.0$. On the AIME24/25/26 average it improves over GRPO by $+4.7$ / $+3.7$ pp on Qwen3-1.7B / 0.6B, and over AdaBack by $+1.0$ / $+1.7$ pp.
  \item \textbf{The ordering MFC $>$ AdaBack $>$ GRPO holds on both base models} on the cross-benchmark average, on Uns-22, and on the AIME average. Mixture and GRPO trade positions across base models. MFC's lead over AdaBack holds on both models, so the advantage of monotone frontier descent (\S\ref{sec:mfc}) is not specific to one model size.
\end{enumerate}

The non-math OOD benchmarks (SciBench, GPQA-Diamond) tell a complementary story: training on $128$ math-only unsolvable problems does not degrade non-math reasoning and shows mild positive transfer for MFC ($+1.2$ to $+4.4$ pp over Base across the two base models), suggesting that the model's improved reasoning capability potentially generalizes beyond mathematics.

\subsection{Reasoning Boundary Expansion}
\label{sec:boundary}

We evaluate whether unsolvable-curriculum training expands the model's reasoning \emph{boundary}.
We measure $\mathrm{pass}@k$ on the union of AIME24/25/26 (90 problems total) for $k \in \{1, 2, 4, 8, 16, 32, 64, 128, 256\}$, using the same five rows of Table~\ref{tab:main} (\emph{Base}, GRPO, Mixture, AdaBack, MFC) on each base model.

Plain GRPO shifts the frontier modestly.
On Qwen3-1.7B, $\mathrm{pass}@64$ rises from $54.4\%$ (Base) to $57.2\%$ ($+2.8$ pp), and $\mathrm{pass}@256$ rises from $57.8\%$ to $60.0\%$ ($+2.2$ pp).
On Qwen3-0.6B, the corresponding gains are $+4.1$ pp at $\mathrm{pass}@64$ and $+3.9$ pp at $\mathrm{pass}@256$.
These gains still trail every teacher-guided method, consistent with prior reports \cite{Yue_et_al.} that standard RL primarily re-weights existing solution paths.
Mixture, AdaBack, and MFC each add several further points at large $k$, and their gains over Base tend to be larger at high $k$.
Large $k$ is the setting that most directly measures reasoning coverage.
MFC is highest on both base models at every $k$.
At $\mathrm{pass}@256$, it reaches $70.0\%$ on Qwen3-1.7B and $53.9\%$ on Qwen3-0.6B, exceeding GRPO by $+10.0$ pp for both models.
Its lead over AdaBack widens at large $k$, reaching $+3.0$ pp at $\mathrm{pass}@256$ for Qwen3-0.6B and $+3.3$ pp for Qwen3-1.7B.
Together with the analysis in \S\ref{sec:dist-shift}, the higher unguided training mass accumulated by MFC is associated with broader reasoning coverage at large $k$.
\subsection{Data Efficiency}
\label{sec:efficiency}

The four algorithms span a $\sim$16$\times$ range in source-data size: $2{,}000$ problems for GRPO, $640$ rows ($128 \times 5$) for Mixture, and $128$ problems for AdaBack and MFC, all under matched compute and matched $1{,}200$ GRPO steps. As summarized in the right panel of Figure~\ref{fig:overview} and Table~\ref{tab:main}, AdaBack and MFC match or exceed the GRPO~2k cross-benchmark average on both base models. Mixture exceeds GRPO on Qwen3-1.7B but trails it on Qwen3-0.6B. Under matched compute, MFC on Uns-128 attains a higher cross-benchmark average than GRPO on OR1-2k while using $\sim$16$\times$ fewer source problems.

\section{Analysis}
\label{sec:analysis}

\subsection{Curriculum Dynamics}
\label{sec:dynamics}

\begin{figure}[t]
\centering
\includegraphics[width=\columnwidth]{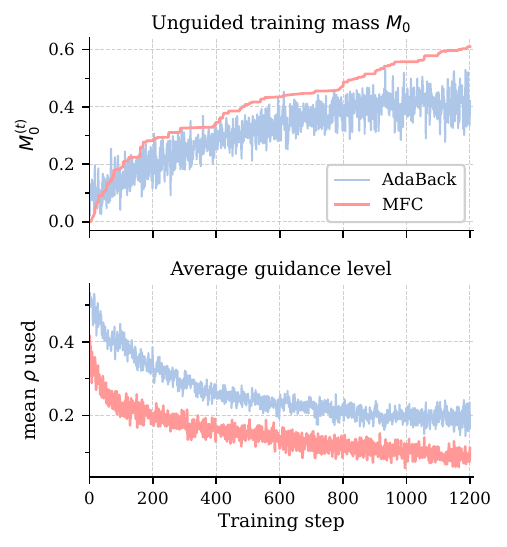}
\caption{Curriculum dynamics of MFC vs.\ AdaBack on Uns-128 (Qwen3-1.7B). Top: unguided training mass $M_0^{(t)}$; bottom: average guidance level used per step. MFC continues to drive training toward the unguided regime throughout the $1{,}200$-step run, whereas AdaBack largely plateaus after step $\sim$800.}
\label{fig:dynamics-compare}
\end{figure}

We now verify that this monotone update raises the unguided training mass $M_0^{(t)}$ (Eq.~\eqref{eq:m0-def}) and lowers the sampled guidance level. Figure~\ref{fig:dynamics-compare} compares MFC and AdaBack on Uns-128 (Qwen3-1.7B) across $M_0^{(t)}$ and $\mathrm{mean\_rho\_used}$, the batch-averaged guidance level $\rho$ actually sampled at each step (lower is closer to the test-time regime). By step $1{,}200$, MFC reaches a smoothed $M_0^{(t)} \approx 0.60$ versus AdaBack's $\approx 0.40$, and drives $\mathrm{mean\_rho\_used}$ down to $\approx 0.09$ versus AdaBack's $\approx 0.20$. The pattern is consistent: MFC's monotone frontier ratchets $\rho_\mathrm{max}$ downward on every successful visit, progressively shifting the sampling distribution toward $g = 0$; AdaBack's bidirectional update stabilizes once a learnable band is established and has no explicit mechanism to push further. The same separation holds on Qwen3-0.6B (Appendix~\ref{sec:appendix-dynamics-06b}).

Appendix~\ref{sec:appendix-gradient} decomposes $\Delta J_0$ into an unguided term weighted by $M_0^{(t)}$ and an augmented term whose sign is unconstrained. MFC's higher $M_0$ corresponds to a larger share of compute spent on prompts with the test-time input format. Empirically, this increase is associated with a higher cross-benchmark average and broader reasoning coverage (Figure~\ref{fig:boundary}).

\subsection{SFT Baseline vs.\ Curriculum RL}
\label{sec:sft-baseline}

A natural concern is that curriculum RL on teacher traces is effectively expensive distillation. We test this with a supervised-fine-tuning (SFT) baseline: Qwen3-1.7B trained on the same Uns-128 problems via next-token prediction on the full teacher trace ($\rho = 1$), using the identical $\rho = 0$ prompt template as the RL evaluation prompt. We train for $50$ epochs (lr $1\mathrm{e}{-5}$, cosine schedule) and select the step-$40$ checkpoint where loss has converged; full configuration in Appendix~\ref{sec:appendix-sft}.

\begin{table}[t]
\centering
\small
\setlength{\tabcolsep}{5pt}
\renewcommand{\arraystretch}{1.1}
\begin{tabular}{lc}
\toprule
\textbf{Method} & \textbf{Avg.\ (9 benchmarks)} \\
\midrule
\textsc{Base}          & 36.8 \\
\textsc{SFT (full trace)} & 39.8 \\
\textsc{AdaBack}       & 45.8 \\
\textbf{MFC} & \textbf{47.3} \\
\bottomrule
\end{tabular}
\caption{Cross-benchmark average $\mathrm{pass}@1$ (\%) on Qwen3-1.7B. SFT uses the same Uns-128 corpus and full teacher traces; curriculum RL methods use the same data under adaptive guidance. Per-benchmark breakdown in Appendix~\ref{sec:appendix-sft}.}
\label{tab:sft-avg}
\end{table}

Table~\ref{tab:sft-avg} shows that SFT improves over Base by only $+3.0$ pp on average, far short of AdaBack ($+9.0$) and MFC ($+10.5$). 
SFT memorizes a single teacher path per problem and converges quickly (under $50$ epochs, see Appendix~\ref{sec:appendix-sft}) on this small corpus; curriculum RL, by contrast, drives $\rho$ to $0$ under reward feedback and forces the policy to discover its own solutions. The curriculum-RL gains are therefore not attributable to teacher-trace distillation: the same $128$ problems and the same teacher information yield qualitatively different outcomes under SFT versus curriculum RL.

\section{Related Work}
\label{sec:related}

\paragraph{Problems beyond the RLVR frontier.}
Studies of reasoning boundaries find that standard RLVR often amplifies solution paths already available to the model, while many problems remain outside its sampled coverage \cite{Yue_et_al.,Havrilla_et_al.}.
Uniform failure creates an optimization problem because a GRPO group with identical rewards has no group-relative advantage.
DAPO dynamic sampling removes groups with accuracy zero or one when constructing a training batch \cite{DAPO}.
RL-ZVP keeps zero-variance groups and supplies an entropy-guided token-level advantage \cite{RL-ZVP}.
SwS identifies weaknesses from persistent training failures and synthesizes new questions around the associated concepts \cite{SwS}.
Adaptive Difficulty Curriculum Learning (ADCL) periodically re-estimates problem difficulty and reorders upcoming batches as model capability changes \cite{Learning_Like_Humans}.
For an unsolvable-only pool, reordering alone does not change uniformly failed rollout groups.
Our setting keeps the original problems selected by $\mathrm{pass}@64=0$.
Partial teacher traces turn each problem into guided states with mixed rollout outcomes, allowing the curriculum to train on the same fixed corpus.

\paragraph{Backward-chaining over partial solutions.}
When a solution trace is available, the amount of revealed reasoning becomes a curriculum variable.
R$^3$ moves the generation starting point through a correct demonstration and mixes examples from several stages \cite{R3}.
In our setup the trace is auxiliary context, and the student generates a complete solution from scratch.
AdaBack adapts the revealed portion per sample through bidirectional updates to a feasible interval \cite{AdaBack}.
SEELE adjusts hint length to keep rollout accuracy near a target level using item response theory \cite{SEELE}.
MFC keeps a single upper frontier and lowers it only after success.
This update is designed to increase unguided training mass when every original prompt is initially infeasible.

\paragraph{Teacher traces and data-efficient RLVR.}
Data-centric studies show that hard problems, high reward variance, and even carefully chosen single examples can be especially useful for RLVR \cite{Hard_Examples,LILO,1-shot_RLVR}.
Our study considers the endpoint where the selected problems have no sampled success before guidance.
Teacher-generated solutions are also widely used as supervised targets or to bootstrap reasoning training \cite{Hinton_et_al._KD,STaR,Reinforced_Self-Training,ReST-MCTS*}.
Here the teacher trace enters the prompt as context for an RL objective, and the student still produces a complete solution.
The SFT comparison in \S\ref{sec:sft-baseline} isolates this use of the same Uns-128 traces as supervised targets.
Appendix~\ref{sec:related-extended} reviews the broader data-selection, curriculum-learning, and teacher-guidance literature.

\section{Conclusion}
\label{sec:conclusion}

We showed that \emph{unsolvable problems}, defined by $\mathrm{pass}@64=0$ for the base model, can be turned into productive RLVR training signal via a stronger teacher's structured trace combined with a backward-chaining curriculum. Training on only $128$ such problems matches or exceeds a $2{,}000$-problem GRPO baseline on the nine-benchmark average for both base models ($\sim$16$\times$ data efficiency) and expands the reasoning boundary ($\mathrm{pass}@k$ gains up to $k = 256$ on AIME24/25/26).

Within this regime we further identified a distribution-shift cost inherent to guided curricula and proposed \textbf{Monotone Frontier Curriculum (MFC)}, whose per-sample frontier descends monotonically under a success-only update. 
MFC outperforms both the GRPO baseline and existing backward-chaining curricula on the cross-benchmark average for both base models.
Our findings suggest that teacher-guided curricula, paired with the right scheduling mechanism, can unlock substantial reasoning gains from a remarkably small pool of hard problems.
\section*{Limitations}
\label{sec:limitations}

Our work relies on a single fixed teacher model (DeepSeek-V3.2) to generate all reasoning traces, and the quality of the resulting curriculum is inevitably tied to the quality of these traces. We do not study how teacher capability interacts with curriculum design. A stronger teacher may raise the attainable $M_0^{(t)}$ per sample, and it may also introduce reasoning patterns whose hints transfer less cleanly to a smaller student.

The framework is trained only on mathematical reasoning with single-turn LLM outputs.
SciBench and GPQA-Diamond provide out-of-domain probes, but they do not constitute a study of scientific reasoning.
Transfer to code generation \cite{AlphaCode} or formal theorem proving is untested.
These domains introduce distinct challenges: code requires multi-file coherence and execution-based verification, while scientific reasoning often lacks the discrete step structure our hint mechanism relies on.
We have not applied the curriculum to \emph{agentic} tasks \cite{SWE-bench,WebArena}, which need sequential environment actions and multi-step tool-use traces.

\bibliography{custom}

\appendix

\section{Extended Related Work}
\label{sec:related-extended}

\paragraph{Data-centric RLVR.}
The choice of training data profoundly impacts RLVR outcomes.
\citet{Hard_Examples} demonstrate that filtering to the hardest problems yields disproportionate gains over uniform sampling, while \citet{LILO} formalize the ``learnability'' criterion showing that maximizing reward variance is equivalent to optimizing expected improvement.
\citet{1-shot_RLVR} push this further, showing a single well-chosen problem can unlock significant reasoning capability.
\citet{Data_Scaling_RLVR} study data efficiency for RLVR through offline subset selection based on diversity, influence, and moderate difficulty, together with online explorability-based rollout pruning, and \citet{CROPI} propose influence-guided data selection to accelerate training.
Complementary work on reward signal design---process reward \cite{Process_Reward,Math-Shepherd} versus outcome reward---also shapes effective data utilization.
Our work identifies the extreme end of this spectrum---problems with \emph{zero} feasibility---and shows they can be made productive through teacher guidance rather than filtering them away.
The treatment of problems beyond the sampled frontier is discussed in \S\ref{sec:related}.

\paragraph{Curriculum learning beyond partial traces.}
Curriculum learning orders or samples training experiences according to difficulty \cite{Curriculum_Learning}.
Reverse curricula in reinforcement learning start near successful states and move the initial state backward as the policy improves \cite{Salimans_&_Chen_2018,RFCL}.
Other curricula act at the batch level by selecting or reordering problems, including schedules that track changes in model capability \cite{Learning_Like_Humans}.
Our guidance lattice acts within each problem by controlling the portion of a trace supplied as context.
Related studies of reasoning-boundary dynamics examine how prolonged RLVR interacts with exploration and coverage \cite{ProRL,yao2025debaterlvrreasoningcapability}.

\paragraph{Teacher traces as targets and as context.}
Knowledge distillation \cite{Hinton_et_al._KD} and imitation learning \cite{DAgger} train students against outputs or actions supplied by a teacher.
Chain-of-thought prompting \cite{Chain-of-Thought} shows that explicit reasoning steps can improve generation, while STaR \cite{STaR} and process reward methods \cite{OmegaPRM} use intermediate reasoning during training.
\citet{DeepSeek-R1} demonstrate that supervised distillation from reasoning traces can produce capable models.
\citet{Reinforced_Self-Training} and \citet{ReST-MCTS*} use generated solutions to bootstrap further training.
STILL-2 distills long reasoning traces and then bootstraps from successful rollouts \cite{STILL-2}.
Our SFT baseline in \S\ref{sec:sft-baseline} compares target-based training with the use of teacher traces as prompt context under curriculum RL.

\section{Implementation Details}
\label{sec:impl}

This appendix specifies the training configuration shared by all RLVR runs in \S\ref{sec:experiments}, the per-method curriculum hyperparameters, and the hardware setup. All training and rollout infrastructure is built on top of verl \cite{verl}; our curriculum algorithms (Mixture, AdaBack, MFC) are implemented as additional curriculum managers within the same verl-based stack, so that the four trained methods share an identical training and inference pipeline.

\subsection{Shared GRPO Training Core}
\label{sec:impl-grpo}

All four trained algorithms (GRPO, Mixture, AdaBack, MFC) on both base models share an identical GRPO training core (Table~\ref{tab:hp-shared}). Two modifications inherited from DAPO \cite{DAPO} are layered on top: an asymmetric PPO clip ratio ($\epsilon_\mathrm{low} = 0.2$, $\epsilon_\mathrm{high} = 0.28$), and a length-based \emph{overlong-response penalty} that linearly penalizes incorrect responses approaching the $8{,}192$-token cap (buffer length $1024$, penalty factor $1.0$). The penalty extends the verifiable reward from $\{0, 1\}$ to $[-1, 1]$ but does not alter the group-relative advantage structure of GRPO. KL regularization is applied as an auxiliary loss term using a low-variance estimator (\texttt{kl\_loss\_type=low\_var\_kl}) rather than mixed into the reward.

\begin{table}[t]
\centering
\footnotesize
\setlength{\tabcolsep}{3pt}
\renewcommand{\arraystretch}{1.1}
\begin{tabular}{@{}p{0.46\columnwidth}p{0.49\columnwidth}@{}}
\toprule
\textbf{Parameter} & \textbf{Value} \\
\midrule
Rollouts per prompt ($n$)    & $8$ \\
Training batch size          & $128$ \\
PPO mini-batch size          & $128$ (full update) \\
Rollout temperature          & $1.0$ (top-$p$ $1.0$) \\
Max prompt length            & $800$ (GRPO~2k) /\newline $2{,}500$ (hint) \\
Max response length          & $8{,}192$ \\
Optimizer                    & AdamW, lr $1\mathrm{e}{-6}$\newline (constant) \\
KL loss coefficient          & $1\mathrm{e}{-3}$,\newline low-var estimator \\
PPO clip ratio (low / high)  & $0.2$ / $0.28$ \\
Entropy coefficient          & $0$ \\
Overlong-response penalty    & enabled, buffer $1024$,\newline factor $1.0$ \\
Advantage normalisation      & by group std (default) \\
Total GRPO steps             & $1{,}200$ \\
Validation interval          & every $10$ steps \\
Validation decoding          & $T{=}0.6$, top-$p{=}0.95$, $n{=}1$ \\
\bottomrule
\end{tabular}
\caption{Shared GRPO training core, identical across all four trained algorithms and both base models in \S\ref{sec:experiments}.}
\label{tab:hp-shared}
\end{table}

\subsection{Per-Method Curriculum Hyperparameters}
\label{sec:impl-curriculum}

Method-specific curriculum hyperparameters are listed in Table~\ref{tab:hp-method}. Mixture has no learnable per-sample state; its only curriculum choice is the discretized $\rho$-level grid. AdaBack follows the public-recipe defaults; two parameters bear brief explanation. The \emph{on-target valve} $p_\mathrm{zero}$ is the small fixed probability of overriding the sampled $\rho$ with $\rho = 0$, introduced in the original AdaBack paper~\cite{AdaBack} to close the train-test distribution mismatch (we use $p_\mathrm{zero} = 0.1$). The \emph{minimum step delta} governs the smallest discrete change to the per-sample hint length per update. MFC reduces to a single hyperparameter, the success-commit threshold $\tau$ (Algorithm~\ref{alg:mfc}).

\paragraph{Discretization of AdaBack.}
We keep AdaBack's hyperparameters in the original continuous-$\rho$ form for fidelity with the public recipe (Table~\ref{tab:hp-method}), but the actual rollout samples on the integer hint lattice so that AdaBack and MFC share an identical sampling space (\S\ref{sec:bc-family}). At each rollout, the continuous bounds $[\rho_\mathrm{min}(s), \rho_\mathrm{max}(s)]$ are mapped to integer bounds via $g = \mathrm{round}(\rho \cdot n_\mathrm{steps}(s))$, an integer $g_\mathrm{used} \sim \mathrm{Uniform}\{g_\mathrm{min}, \ldots, g_\mathrm{max}\}$ is drawn, and the bidirectional update is applied directly on the integer interval (the minimum step delta of $1$ ensures each successful update strictly contracts it). This rules out the round-induced sampling bias at $g = 0$ that would otherwise systematically penalize AdaBack's measured $M_0^{(t)}$ relative to MFC's discrete-uniform draw.

\begin{table}[t]
\centering
\footnotesize
\setlength{\tabcolsep}{3pt}
\renewcommand{\arraystretch}{1.1}
\begin{tabular}{@{}lp{0.34\columnwidth}p{0.32\columnwidth}@{}}
\toprule
\textbf{Method} & \textbf{Parameter} & \textbf{Value} \\
\midrule
\textsc{GRPO}    & training corpus                   & OR1-2k\newline ($2{,}000$ problems) \\
                 & teacher guidance                  & none ($\rho \equiv 0$) \\
\midrule
\textsc{Mixture} & training corpus                   & Uns-128 expansion\newline ($640$ rows) \\
                 & $\rho$-level grid                 & $\{0, 0.25, 0.5,$\newline $0.75, 1\}$ \\
\midrule
\textsc{AdaBack} & training corpus                   & Uns-128 \\
                 & success threshold $\tau$          & $0.5$ \\
                 & first-query $\rho$                 & $0.5$ \\
                 & on-target valve $p_\mathrm{zero}$ & $0.1$ \\
                 & minimum step delta                & $1$ \\
\midrule
\textbf{MFC}     & training corpus                   & Uns-128 \\
                 & success threshold $\tau$          & $\mathbf{0.5}$ \\
                 & initial $\rho_\mathrm{max}$       & $1.0$ \\
\bottomrule
\end{tabular}
\caption{Method-specific curriculum hyperparameters. The three teacher-guided methods all use hint-mode guidance with the discretization of \S\ref{sec:bc-family}.}
\label{tab:hp-method}
\end{table}

\subsection{Hardware and Wall-Clock}
\label{sec:impl-hardware}

All RLVR runs use $8 \times$ NVIDIA RTX PRO 6000 (96~GB) GPUs; FSDP \cite{FSDP} with parameter and optimizer sharding is used throughout (offload disabled). Rollouts are served by vLLM \cite{vLLM} at $70\%$ GPU memory utilization with prefix caching enabled. Dynamic batching uses \texttt{ppo\_max\_token\_len\_per\_gpu} $= 48{,}000$ and \texttt{ppo\_micro\_batch\_size\_per\_gpu} $= 4$. A single $1{,}200$-step training run takes approximately $90$ wall-clock hours for Qwen3-1.7B and approximately $70$ hours for Qwen3-0.6B on the same $8 \times$ RTX PRO 6000 setup. Including all diagnostic and development runs, our total compute usage amounted to approximately $\mathbf{7{,}000}$ RTX PRO 6000 GPU hours.

\section{Training-Data Construction}
\label{sec:appendix-data}

This appendix details the construction pipeline for the four corpora used in \S\ref{sec:setup}: the mixed-difficulty training set \textbf{OR1-2k}, the held-out test set \textbf{OR1-200}, the $128$-problem unsolvable training set \textbf{Uns-128}, and the held-out unsolvable subset \textbf{Uns-22}. All four are derived from the \emph{train} split of OpenR1-Math-220k \cite{OpenR1-Math} ($93.7$k problems) under a single, deterministic protocol with random seed $42$.

\paragraph{Filtering and train/test split ($\to$ OR1-2k, OR1-200).}
Starting from the $93.7$k-problem train split, we apply a chain of quality filters to keep the corpus narrowly scoped to verifiable, single-answer competition-level mathematical reasoning problems (substantially harder than elementary benchmarks such as GSM8K \cite{GSM8K}):
\begin{itemize}
  \item drop problems where the released DeepSeek-R1 reference solution is judged incorrect by Math-Verify\footnote{\url{https://github.com/huggingface/Math-Verify}};
  \item drop multiple-choice problems (where the verifiable reward degenerates to chance);
  \item drop problems whose statement is too short or too long, and problems whose released reference solution is too long, to keep prompt and rollout budgets bounded;
  \item drop problems whose ground-truth answer string contains a comma (a strong indicator of multi-answer or list-valued problems that defeat exact-match scoring).
\end{itemize}
From the surviving pool we randomly draw $2{,}200$ problems under seed $42$ and split them into $2{,}000$ training problems (\textbf{OR1-2k}) and $200$ held-out test problems (\textbf{OR1-200}), keeping the splits disjoint. Difficulty composition is preserved by construction (random sampling), and the resulting train set spans the full $\mathrm{pass}@64$ spectrum from trivial ($\mathrm{pass}@64 \approx 1$) to unsolvable ($\mathrm{pass}@64 = 0$).

\paragraph{Unsolvable filtering ($\to$ Uns-128).}
The unsolvable pool is derived only from OR1-2k:
\begin{enumerate}
  \item Re-evaluate every problem in OR1-2k with Qwen3-1.7B at the response-length cap used at training time ($8{,}192$ tokens) and retain those with $\mathrm{pass}@64 = 0$ ($147$ problems, $\sim$7\% of the train set).
  \item Generate a structured reasoning trace for every retained problem with the DeepSeek-V3.2 \cite{DeepSeek-V3.2} teacher (the prompt template, structured-output schema, and step-extraction rule are described in Appendix~\ref{sec:prompt-templates}). Drop any problem on which the teacher itself fails the Math-Verify check, leaving $136$ raw rows.
  \item Sort the surviving problems by combined question-plus-trace character length and take the $128$ shortest. This length-controlled selection bounds the maximum prompt length seen at training time and removes a small fraction of pathologically long traces; it is otherwise content-agnostic.
\end{enumerate}
The resulting $128$-problem set is \textbf{Uns-128}, the single unsolvable training corpus shared by Mixture, AdaBack, and MFC.

\paragraph{In-domain unsolvable test set ($\to$ Uns-22).}
For evaluation purposes, we additionally extract the $22$-problem subset of OR1-200 with $\mathrm{pass}@64 = 0$ on Qwen3-1.7B, using the same $8{,}192$-token response cap as the unsolvable filter above. This subset, \textbf{Uns-22}, is a strict held-out evaluation set: it shares no problems with Uns-128 and is never used for training or hyperparameter selection.

\paragraph{Reuse for Qwen3-0.6B.}
Uns-128 was filtered using Qwen3-1.7B's $\mathrm{pass}@64$ but is reused unchanged when training Qwen3-0.6B. Because Qwen3-0.6B is a strictly weaker sibling from the same model family and training generation, every problem unsolvable for Qwen3-1.7B is at least as hard for Qwen3-0.6B. Sharing the unsolvable corpus across base models keeps the comparison apples-to-apples and avoids a confound where the smaller model would otherwise see an easier curriculum.

\paragraph{Data release.}
We release complete records for OR1-2k, OR1-200, Uns-128, and Uns-22 as a public Hugging Face dataset repository.\footnote{\url{https://huggingface.co/datasets/yukangzhu/unlocking-the-unsolvable}}
The four configs \texttt{or1\_2k}, \texttt{uns128}, \texttt{or1\_200}, and \texttt{uns22} can be loaded and used directly.
Teacher traces appear only in the Uns-128 config.
These files can be used for training or evaluation without mapping their identifiers back to OpenR1-Math-220k.

\section{Additional Experimental Details}
\label{sec:extra}

\subsection{Curriculum Metric Definitions}
\label{sec:appendix-metrics}

Throughout \S\ref{sec:dynamics} we report two curriculum-level metrics, computed per training step:

\begin{itemize}
  \item \textbf{Unguided training mass} $M_0^{(t)}$ (Eq.~\eqref{eq:m0-def}; logged as \texttt{frac\_effective\_zero} or \texttt{frac\_at\_zero}): the fraction of in-batch prompt visits whose discretized hint length $g(s, \rho) = 0$. Higher values indicate that a larger share of training compute is spent on prompts matching the test-time distribution.
  \item \textbf{Mean guidance level} $\mathrm{mean\_rho\_used}$: the batch-averaged value of the guidance level $\rho$ actually sampled at each step, i.e.\ $\frac{1}{|\mathcal{B}|} \sum_{(s,\rho) \in \mathcal{B}} \rho$. Lower values indicate the curriculum is closer to the unguided regime on average.
\end{itemize}

\subsection{Qwen3-0.6B Curriculum Dynamics}
\label{sec:appendix-dynamics-06b}

\begin{figure*}[t]
\centering
\includegraphics[width=\textwidth]{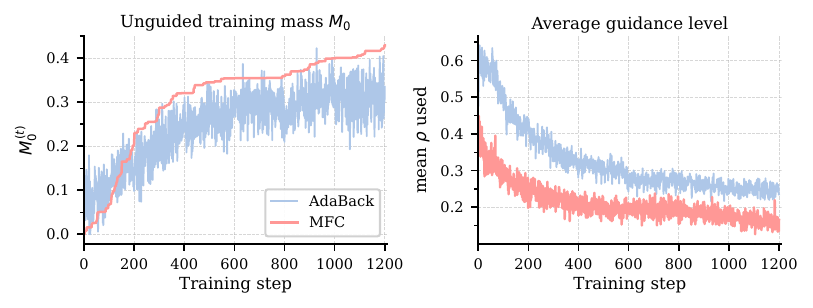}
\caption{Curriculum dynamics of MFC vs.\ AdaBack on Uns-128 (Qwen3-0.6B). Left: unguided training mass $M_0^{(t)}$; right: average guidance level. The qualitative separation is identical to the 1.7B results (Figure~\ref{fig:dynamics-compare}).}
\label{fig:dynamics-06b}
\end{figure*}

Figure~\ref{fig:dynamics-06b} replicates the curriculum-dynamics comparison of Figure~\ref{fig:dynamics-compare} (main text, Qwen3-1.7B) on Qwen3-0.6B. The qualitative pattern is identical: MFC's $M_0^{(t)}$ continues to climb throughout training while AdaBack plateaus, and MFC's $\mathrm{mean\_rho\_used}$ descends well below AdaBack's. This confirms that MFC's structural advantage is not specific to the 1.7B scale.

\subsection{Training and Validation Reward Dynamics}
\label{sec:appendix-reward-dynamics}

\begin{figure*}[t]
\centering
\includegraphics[width=\textwidth]{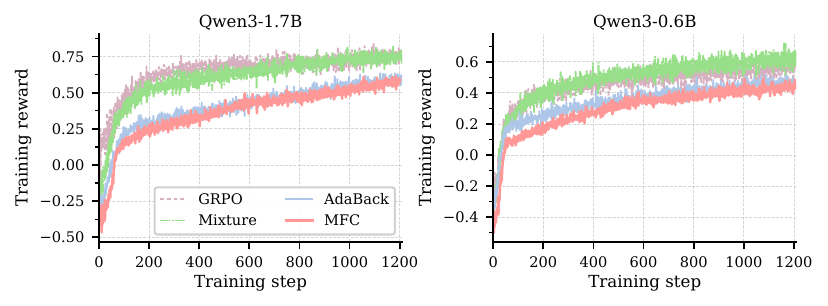}
\caption{Training reward (mean per batch) across $1{,}200$ steps for all four methods on both base models. Note that training reward is not directly comparable across methods because curriculum mechanisms present samples at different effective difficulty levels; see text for interpretation.}
\label{fig:train-reward}
\end{figure*}

\begin{figure*}[t]
\centering
\includegraphics[width=\textwidth]{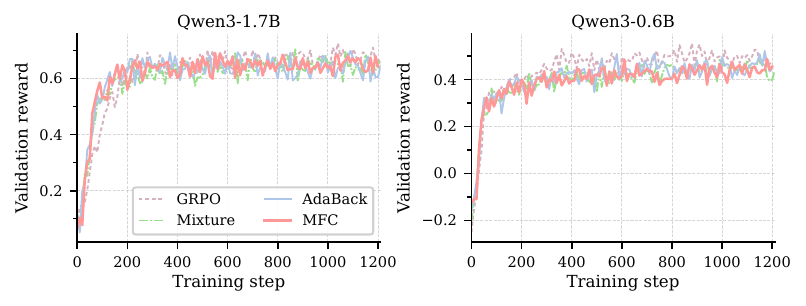}
\caption{Validation reward on OR1-200 (no guidance, $\rho = 0$) evaluated every $10$ steps. This is the fair comparison of unguided capability across methods. All methods converge by step $1{,}200$; the curriculum methods (especially MFC) rise faster in the first $\sim$400 steps.}
\label{fig:val-reward}
\end{figure*}

Figures~\ref{fig:train-reward} and~\ref{fig:val-reward} show the training reward and validation reward across all four methods on both base models.

\paragraph{Training reward (Figure~\ref{fig:train-reward}).}
An important caveat: training reward is \emph{not} directly comparable across methods. The curriculum mechanisms (Mixture, AdaBack, MFC) present each sample at a guidance level selected by their respective scheduling rules; higher $\rho$ yields easier effective prompts and thus higher expected reward. GRPO, by contrast, trains on mixed-difficulty data without guidance. Consequently, the absolute level and the trajectory of training reward reflect the effective difficulty of the curriculum at each step, not the model's test-time capability.

\paragraph{Validation reward (Figure~\ref{fig:val-reward}).}
Validation is always performed on OR1-200 without any teacher guidance ($\rho = 0$), making it a fair apples-to-apples comparison of unguided capability at each checkpoint. All four methods converge to broadly similar validation reward by step $1{,}200$, but the curriculum methods reach this level faster (fewer wasted early steps) and MFC shows the steepest early rise, consistent with its higher $M_0^{(t)}$ allocating more on-target gradient compute from the start.

\section{SFT Baseline Details}
\label{sec:appendix-sft}

This appendix provides full details for the SFT baseline of \S\ref{sec:sft-baseline}.

\paragraph{Data construction.}
Each of the $128$ Uns-128 samples is formatted as a single-turn model conversation: the system and user messages are identical to the $\rho = 0$ prompt template used during curriculum RL evaluation, with no hints or guidance of any kind; the assistant message contains the full teacher reasoning wrapped in \texttt{<think>}$\ldots$\texttt{</think>} tags followed by \texttt{\textbackslash boxed\{teacher\_answer\}}, exactly matching the generation format trained by curriculum RL. The \texttt{teacher\_answer} field is recovered from the original training-pool metadata to ensure complete fidelity to the teacher model's own output.

\paragraph{Training configuration.}
We train Qwen3-1.7B (base) via full-parameter SFT (no LoRA) using FSDP on $2\times$ NVIDIA RTX PRO 6000 GPUs. Key hyperparameters: global batch size $128$ (entire dataset in one step), learning rate $1\mathrm{e}{-5}$ with cosine decay and $10\%$ warmup, weight decay $0.01$, gradient clipping at $1.0$, bf16 mixed precision, max sequence length $2{,}048$ tokens, $50$ epochs ($= 50$ gradient steps). The $2{,}048$-token cap is shorter than the $8{,}192$-token response cap used by curriculum RL, but exceeds the full prompt-plus-trace length of every problem in Uns-128, so no truncation occurs in the SFT data. Total wall-clock time is approximately $15$ minutes.

\begin{figure}[t]
\centering
\includegraphics[width=\linewidth]{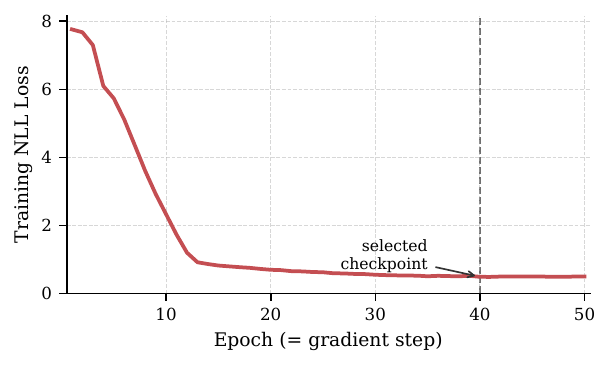}
\caption{SFT training NLL loss over $50$ epochs on Uns-128 (Qwen3-1.7B). The dashed line marks the selected checkpoint at step $40$ where loss has effectively converged.}
\label{fig:sft-loss}
\end{figure}

\paragraph{Checkpoint selection.}
Figure~\ref{fig:sft-loss} shows the training loss curve. Since the validation set equals the training set for this $128$-sample corpus, we show only training NLL. Loss levels off by step ${\sim}35$--$40$. We select step $40$ as the evaluation checkpoint.

\paragraph{Per-benchmark results.}
Table~\ref{tab:sft-full} reports the full per-benchmark $\mathrm{pass}@1$ for the SFT baseline (Qwen3-1.7B, step $40$).

\begin{table*}[t]
\centering
\small
\setlength{\tabcolsep}{4.5pt}
\renewcommand{\arraystretch}{1.1}
\begin{tabular}{lcccccccccc}
\toprule
\textbf{Method}
  & \textbf{OR1-200} & \textbf{Uns-22} & \textbf{MATH-500} & \textbf{AMC23}
  & \textbf{AIME24} & \textbf{AIME25} & \textbf{AIME26}
  & \textbf{SciBench} & \textbf{GPQA-D}
  & \textbf{Avg.} \\
\midrule
\textsc{SFT} & 54.5 & 13.6 & 83.4 & 61.9 & 26.7 & 26.7 & 19.2 & 36.0 & 36.5 & 39.8 \\
\bottomrule
\end{tabular}
\caption{Per-benchmark $\mathrm{pass}@1$ (\%) for the SFT baseline on Qwen3-1.7B. Compare with Base ($36.8$ avg) and MFC ($47.3$ avg) in Table~\ref{tab:main}.}
\label{tab:sft-full}
\end{table*}

\section{Formal View of the Unguided Training Mass Surrogate}
\label{sec:appendix-gradient}

This appendix formalizes the on-target / augmented gradient decomposition that motivates the unguided training mass $M_0^{(t)}$ of Eq.~\eqref{eq:m0-def} as the surrogate of choice in \S\ref{sec:dist-shift}, and gives a careful comparison of MFC and AdaBack on $M_0^{(t)}$.

\paragraph{First-order Taylor decomposition.}
Let $\theta$ denote the current parameters and consider one gradient step against the curriculum objective $J_\mathcal{C}^{(t)}$ of Eq.~\eqref{eq:train-obj}, $\theta' = \theta + \eta \nabla J_\mathcal{C}^{(t)}(\theta)$ for a small step size $\eta > 0$. The first-order Taylor expansion of the test-time objective $J_0$ of Eq.~\eqref{eq:test-obj} gives
\begin{equation}
\label{eq:appendix-taylor}
\begin{aligned}
J_0(\theta') - J_0(\theta) \;=\;& \eta\,\bigl\langle \nabla J_0(\theta),\, \nabla J_\mathcal{C}^{(t)}(\theta) \bigr\rangle \\
&+\; O(\eta^2).
\end{aligned}
\end{equation}
Decompose the curriculum gradient on the on-target ($g = 0$) / augmented ($g > 0$) partition,
\begin{equation}
\label{eq:appendix-decomp}
\begin{aligned}
\nabla J_\mathcal{C}^{(t)}(\theta) \;=\;& M_0^{(t)}\,\nabla J_\mathrm{zero}(\theta) \\
&+\, \bigl(1 - M_0^{(t)}\bigr)\,\nabla J_\mathrm{aug}(\theta),
\end{aligned}
\end{equation}
where $\nabla J_\mathrm{zero}$ is the gradient of the conditional curriculum objective restricted to visits with $g(s, \rho) = 0$, $\nabla J_\mathrm{aug}$ the analogous quantity restricted to $g > 0$, and $M_0^{(t)}$ as in Eq.~\eqref{eq:m0-def}. Substituting Eq.~\eqref{eq:appendix-decomp} into Eq.~\eqref{eq:appendix-taylor},
\begin{equation}
\label{eq:appendix-final}
\begin{aligned}
\Delta J_0 \;=\;& \eta\,M_0^{(t)}\,\bigl\langle \nabla J_0,\,\nabla J_\mathrm{zero} \bigr\rangle \\
&+\, \eta\,\bigl(1 - M_0^{(t)}\bigr)\,\bigl\langle \nabla J_0,\,\nabla J_\mathrm{aug} \bigr\rangle \\
&+\, O(\eta^2).
\end{aligned}
\end{equation}

\paragraph{The on-target term under unbiased sampling.}
Under the assumption that the conditional distribution of $s$ given $g(s, \rho) = 0$ matches $\mathcal{D}$, $\nabla J_\mathrm{zero}$ is an unbiased estimator of $\nabla J_0$ in expectation, and the first term of Eq.~\eqref{eq:appendix-final} satisfies
\begin{equation*}
\mathbb{E}\bigl[\bigl\langle \nabla J_0,\,\nabla J_\mathrm{zero} \bigr\rangle\bigr] \;=\; \|\nabla J_0\|^2 \;\geq\; 0,
\end{equation*}
contributing a definite-sign improvement to $J_0$ at an expected rate proportional to $M_0^{(t)}$.
This condition holds for Mixture's $\rho = 0$ row and for the portion of AdaBack visits induced by its fixed $p_\mathrm{zero}$ valve.
It does not hold in general for MFC, since its lattice-floor probability $1/(g_\mathrm{curr}(s)+1)$ depends on the per-sample frontier.
For MFC, $M_0^{(t)}$ remains the weight on unguided visits with the test-time input format, but those visits follow a frontier-reweighted distribution over problems rather than $\mathcal{D}$.

\paragraph{The augmented term has unconstrained sign.}
The second term of Eq.~\eqref{eq:appendix-final} is mediated entirely through parameter sharing across distinct prompts. Its sign and magnitude depend on whether the gradient on augmented prompts $s_\rho$ is well-aligned with the gradient on the unguided prompt $s$, which is not in general guaranteed and can be:
\begin{itemize}
  \item positive (helpful generalization, when reasoning learned under partial guidance transfers to unaided solving);
  \item zero (neutral, when the model learns hint-conditioned behaviors that do not transfer);
  \item negative (negative transfer, when the model overfits to teacher-style scaffolding or prompt artifacts that hurt unguided generation).
\end{itemize}
Cauchy--Schwarz bounds the magnitude by $\|\nabla J_\mathrm{aug}\|\,\|\nabla J_0\|$, but does not pin the sign.

\paragraph{Implication for surrogate design.}
Eq.~\eqref{eq:appendix-final} suggests that a curriculum cannot expect to improve $J_0$ purely through augmented visits, but neither can it improve $J_0$ by visiting only $\rho = 0$ when those visits produce uniform failure (in which case $\|\nabla J_\mathrm{zero}\|$ itself collapses by the same advantage-normalization mechanism that makes unsolvable problems inert under standard GRPO). Both factors of the on-target term, $M_0^{(t)}$ and $\|\nabla J_\mathrm{zero}\|$, must be advanced jointly. MFC's per-sample frontier monotonically raises the probability of an unguided visit after a successful visit at a lower $g$ (Algorithm~\ref{alg:mfc}). The value $g = 0$ remains in the sampling support from the first visit, although early unguided groups may still yield vanishing gradients before unaided success becomes likely.

\paragraph{Comparison with AdaBack on $M_0^{(t)}$.}
Conditional on the per-sample state, MFC's on-target visit probability is exactly $1/(g_\mathrm{curr}(s) + 1)$ by the discrete-uniform draw on $\{0, 1, \ldots, g_\mathrm{curr}(s)\}$; since $g_\mathrm{curr}$ is monotone non-increasing under MFC's success-only update, this conditional probability is monotone non-decreasing across visits to $s$. AdaBack admits no such monotone bound: under the same discrete-uniform sampling (\S\ref{sec:bc-family}), its per-sample on-target visit probability is $p_\mathrm{zero} + (1 - p_\mathrm{zero})\,\mathbf{1}[g_\mathrm{min}(s) = 0]\,/\,(g_\mathrm{max}(s) - g_\mathrm{min}(s) + 1)$ because the valve overrides the uniform draw. When $g_\mathrm{min}(s) > 0$, the uniform-draw contribution vanishes and the on-target probability equals $p_\mathrm{zero}$. A subsequent success at any $g$ resets $g_\mathrm{min}(s)$ to $0$, restoring uniform mass at $g = 0$. In our discrete unsolvable-only runs this mass is intermittent, and AdaBack's aggregate $M_0^{(t)}$ rises before plateauing around $0.40$ (Figure~\ref{fig:adaback-m0}).

\section{Prompt Templates}
\label{sec:prompt-templates}

This section provides the prompt templates used for teacher trace generation and hint-based guidance throughout the paper.

\subsection{Teacher Trace Generation}
\label{sec:prompt-teacher}

To obtain structured reasoning traces for each unsolvable problem, we prompt DeepSeek-V3.2 with the following system prompt. The output is required to consist of discrete \texttt{<step>} tags wrapped inside a \texttt{<solution>} block, followed by a single \texttt{<answer>} tag. We parse the output by extracting the content of each \texttt{<step>} tag into an ordered list of reasoning steps.

\begin{center}
\begin{tcolorbox}[
    enhanced,
    breakable,
    colback=gray!3,
    colframe=gray!60,
    coltitle=black,
    colbacktitle=gray!15,
    fonttitle=\bfseries,
    width=1.0\columnwidth,
    title=System Prompt for Teacher Trace Generation
]
\small
\texttt{\# [Persona Definition]}

You are an expert mathematics teacher preparing educational material for students. Your goal is to explain the solution to a complex math problem in a way that is clear, logical, and easy to follow. Imagine your student is intelligent but may miss intermediate steps if you are not explicit.

\medskip
\texttt{\# [Strict Formatting Rules \& Few-Shot Example]}

Your entire response MUST be contained within a single \texttt{<solution>} tag. You MUST reason step by step. You MUST wrap each logical reasoning step in a \texttt{<step>...\allowbreak</step>} tag. You MUST wrap ONLY the final answer in an \texttt{<answer>...\allowbreak</answer>} tag. Do not include any additional explanations, comments, or text inside the \texttt{<answer>} tag. Do not output any text outside these tags.

\medskip
\texttt{Example:}

\medskip
\textbf{User:} The real roots of the equations $x^{5}+x+1=0$ and $x+\sqrt[5]{x}+1=0$ are $\alpha, \beta$ respectively, then $\alpha+\beta$ equals?

\medskip
\textbf{Assistant:}\\
\texttt{<solution>}\\
\texttt{<step>}Let's define a function based on the first equation: $f(x) = x^5 + x + 1$. Its derivative is $f'(x) = 5x^4 + 1$, which is always positive.\texttt{</step>}\\
\texttt{<step>}The first equation has real root $\alpha$, so $f(\alpha) = 0$.\texttt{</step>}\\
\texttt{<step>}From the second equation, $\sqrt[5]{\beta} = -\beta - 1$.\texttt{</step>}\\
\texttt{<step>}Evaluating $f(-1-\beta) = (-1-\beta)^5 + (-1-\beta) + 1 = \beta - 1 - \beta + 1 = 0$.\texttt{</step>}\\
\texttt{<step>}Since $f$ is strictly increasing, $\alpha = -1-\beta$, so $\alpha+\beta=-1$.\texttt{</step>}\\
\texttt{<answer>}$-1$\texttt{</answer>}\\
\texttt{</solution>}
\end{tcolorbox}
\end{center}

\subsection{Hint-Based Guidance Prompt}
\label{sec:prompt-hint}

During training, a guidance level $\rho$ determines how many of the teacher's extracted steps are revealed to the student (Eq.~\ref{eq:rho-lattice}). The selected steps are injected into the user message as a hint. The full prompt seen by the student model consists of a system message and a user message constructed as follows:

\begin{center}
\begin{tcolorbox}[
    enhanced,
    breakable,
    colback=gray!3,
    colframe=gray!60,
    coltitle=black,
    colbacktitle=gray!15,
    fonttitle=\bfseries,
    width=1.0\columnwidth,
    title=System Prompt (Student Model)
]
\small
You are an expert mathematician with strong problem-solving skills. Think step by step.
\end{tcolorbox}
\end{center}

\begin{center}
\begin{tcolorbox}[
    enhanced,
    breakable,
    colback=gray!3,
    colframe=gray!60,
    coltitle=black,
    colbacktitle=gray!15,
    fonttitle=\bfseries,
    width=1.0\columnwidth,
    title=User Message Template (with guidance)
]
\small
\texttt{\{question\}}

\medskip
Below are some initial reasoning steps that may help you:

\texttt{\{guidance\_steps\}}

\medskip
Please solve the problem step by step. You should use the provided steps as a reference, but do NOT just copy them. Instead, reconstruct the complete reasoning process in your own words, starting from the beginning, and continue the reasoning to find the final answer.

\medskip
Use this format:\\
\texttt{<think>}\\
{[Your reasoning process here]}\\
\texttt{</think>}\\
\texttt{\textbackslash boxed\{answer\}}
\end{tcolorbox}
\end{center}

When $\rho = 0$ (no guidance), the hint block is omitted and the user message reduces to the question followed by the standard solving instruction.

\section{Broader Impacts}
\label{sec:broader-impacts}

This work advances methods for improving mathematical reasoning in LLMs via reinforcement learning. The potential negative societal impacts align with those generally associated with LLM reasoning technologies: stronger reasoning capabilities could in principle be misused to generate more convincing misinformation or to automate harmful planning tasks. However, our contribution is methodological (a curriculum scheduling algorithm) and operates on publicly available base models; it does not introduce new capabilities beyond what larger-scale training on public data already enables. On the positive side, our data-efficiency finding (${\sim}16\times$ reduction in required training data) may help democratize access to reasoning improvements for resource-constrained research groups.

\section{Licenses of Artifacts Used}
\label{sec:licenses}

\paragraph{Models.}
Qwen3-1.7B and Qwen3-0.6B (base models) are downloaded from their official HuggingFace repositories, where they are released under the Apache 2.0 license.
DeepSeek-V3.2 (teacher model) was accessed exclusively through the OpenRouter API for the purpose of generating teacher reasoning traces; use of the API and the resulting outputs is governed by the provider's terms of service. We do not redistribute the model weights or any portion of them.

\paragraph{Training data.}
OpenR1-Math-220k is released under the Apache 2.0 license.

\paragraph{Evaluation benchmarks.}
MATH-500 is a subset of the MATH dataset; the source dataset and its splits are released under the MIT license.
SciBench is released under the MIT license.
GPQA-Diamond is a subset of GPQA, whose source dataset is released under CC-BY 4.0; the specific redistribution we use does not specify a separate license.
AMC23 and AIME24/25/26 are mathematics competition problems copyrighted by the Mathematical Association of America (MAA); we access them through community-redistributed datasets on HuggingFace and use them solely for non-commercial research evaluation, consistent with standard practice in the LLM reasoning literature.

\paragraph{Infrastructure.}
verl (training framework), vLLM (inference engine), and Math-Verify (answer verification) are all released under the Apache 2.0 license.

All artifacts are used in accordance with their stated terms.
We release the training and in-domain evaluation datasets used in this work, together with the Uns-128 teacher traces, in a public Hugging Face dataset repository.\footnote{\url{https://huggingface.co/datasets/yukangzhu/unlocking-the-unsolvable}}
The repository documents the source of each record, the construction procedure, and the model used to generate the traces.
The released bundle is distributed under the Apache 2.0 license, matching the Apache 2.0 license of its OpenR1-Math-220k source; we apply the same license to our own rights, if any, in the DeepSeek-V3.2 teacher traces, whose generation remains governed by the provider's terms of service.
Because the upstream problems are drawn from olympiad, competition, and examination sources, a dataset-level license does not by itself clear rights in every underlying problem, and users remain responsible for those rights.
We do not release model weights.

\section{Use of AI Assistants}
\label{sec:ai-use}

Large language models (Claude, ChatGPT) were used to assist with grammar checking, paraphrasing, and improving clarity of the authors' original content.

\end{document}